\documentclass[runningheads]{llncs}
\usepackage[T1]{fontenc}

\usepackage{graphicx}

\usepackage{graphicx}
\usepackage{amsmath}
\usepackage{amssymb}
\usepackage{booktabs}
\usepackage{colortbl}
\usepackage{float}
\usepackage{multirow}
\usepackage{array}
\usepackage{placeins}
\usepackage{xcolor}

\usepackage{hyperref}

\usepackage[style=numeric, sorting=none]{biblatex}
\begin{document}
\title{Bayesian adaptively-weighted ensembles for few-shot abdominal segmentation}
\titlerunning{Bayesian ensembles for few-shot segmentation}

\author{Abbas Al-Sabbagh\inst{1,\dagger} \and
Shalom F. Mushtaq\inst{1,\dagger} \and
Tomás M. da Silva\inst{1,\dagger}
\and
Kushagra Soni \inst{1,\dagger} \and
Binawei Gbamila \inst{1,2} \and
Sri Atluri \inst{1,2} \and
Qianye Yang \inst{4} \and
Yipeng Hu \inst{3,4} \and
Claire C. Villette \inst{1,2} \and
Shaheer U. Saeed \inst{1,2,3,*}
}
\authorrunning{A. Al-Sabbagh et al.}

\institute{Centre for Bioengineering, School of Engineering and Materials Science, Queen Mary University of London, London, United Kingdom
\and
Digital Environment Research Institute, Queen Mary University of London, London, United Kingdom\\
\and
UCL Hawkes Institute; Department of Medical Physics and Biomedical Engineering, University College London, London, United Kingdom\\
\and
Institute of Biomedical Engineering, Department of Engineering Science, University of Oxford, Oxford, United Kingdom\\
Note $\dagger$: Contributed equally\\
*Email: \email{shaheer.saeed@qmul.ac.uk}}

\maketitle

\begin{abstract}
Few-shot learning has emerged as a promising approach for anatomical segmentation when labelled data are scarce. However, different few-shot learning algorithms exhibit complementary strengths and weaknesses, with performance varying across anatomical targets and institutions. Existing few-shot segmentation ensembles, that combine predictions from multiple algorithms, typically employ fixed weighting schemes and therefore cannot adjust model contributions according to the target domain. In this work, we propose a Bayesian adaptively-weighted ensemble framework for segmentation under label scarcity and domain shift. Multiple few-shot segmentation algorithms are first adapted using a small labelled support set. Bayesian optimisation is then used to automatically identify ensemble weights that maximise segmentation performance on a target-domain validation set. The learned weights are subsequently  fixed and applied to combine predictions on previously unseen query images from the target domain. The proposed framework is evaluated on the Cross-institution Male Pelvic Structures dataset using held-out anatomical structures and institutions to simulate simultaneous label scarcity and institutional domain shift. Results demonstrate statistically significant improvements over individual few-shot learners, fixed-weight ensembles, training-from-scratch baselines and recent state-of-the-art ensembling approaches. By adapting model contributions to the target anatomy and institutional domain, the proposed framework provides a practical mechanism for deploying segmentation systems to new clinical sites under severe annotation constraints. Code: \url{github.com/kushuu/few-shot-3d-image-segmentation}

\end{abstract}

\section{Introduction}

Anatomical segmentation is a critical prerequisite for magnetic resonance (MR) image analysis, enabling accurate diagnosis, treatment planning, and image-guided interventions \cite{Chen2020}. For anatomical structures with abundant labelled data, deep neural networks such as U-Net and its variants achieve accurate and reproducible delineations \cite{Ronneberger2015,Oktay2018,Chen2024,Isensee2021,Milletari2016}. However, performance deteriorates for specialised structures that are infrequently annotated, such as the neurovascular bundle \cite{meyer2021domain,Zhang2022}, where limited training data restricts model generalisation \cite{PACHETTI2024102949}. Acquiring additional annotations for these targets is expensive, time-consuming, and subject to inter-observer variability \cite{SAEED2024103181, 10.1007/978-3-030-78191-0_55}. In abdominal magnetic resonance imaging (MRI), this challenge is further compounded by differences in acquisition protocols, scanner characteristics, annotation practices, and patient populations across institutions, alongside substantial inter-patient anatomical variability \cite{li2023prototypical, saeed2022image, chalcroft2021development}. Consequently, accurate segmentation in data-constrained settings remains an open problem.

Few-shot segmentation seeks to address this limitation by adapting models to new anatomical targets using only a small number of labelled examples \cite{Dissanayake2025}. Existing approaches include transfer learning through fine-tuning \cite{PACHETTI2024102949}, gradient-based meta-learning \cite{Finn2017, nichol2018first}, and prototype-based methods \cite{Wang2019PANetFI, li2023prototypical}. While effective in certain settings, these approaches exhibit complementary strengths and weaknesses. Fine-tuning methods adapt a pre-trained neural network using limited examples of the target anatomy. Performance of such approaches is dictated largely by the size of the target anatomy support set \cite{Dissanayake2025}. 
Gradient-based meta-learning methods learn from various tasks in meta-training through soft aggregated gradient updates and are designed for rapid adaptation to target tasks. However, their performance depends heavily on how closely the target anatomy resembles the range of tasks used during meta-training \cite{Finn2017}. Prototype-based methods rely on support-derived feature representations, making them sensitive to support-set quality, anatomical variability, and image contrast \cite{Wang2019PANetFI}. As a result, different few-shot methods may perform better for different anatomical targets and may generalise differently across institutions due to variations in acquisition protocols, image appearance, and patient populations. These complementary characteristics suggest that combining multiple few-shot approaches may yield more robust segmentation performance in data-constrained settings.

Ensembling offers a potential solution by combining complementary model predictions. While ensemble methods have demonstrated benefits in classification and detection tasks \cite{TEKIN2025, LIN2022}, extending them to segmentation is challenging due to the need to aggregate dense voxel-level predictions. Recent segmentation ensembles typically employ fixed averaging schemes or manually specified weights \cite{GANAIE2022105151, DANG2024, khoong2020busu}. However, these fixed equal-weight or randomly assigned ensembles cannot adapt to differences in algorithm reliability across target tasks or institutions. In few-shot segmentation, where both anatomical targets and institutional domains may favour different algorithms, ensemble weights should be adapted to the target deployment setting \cite{Dissanayake2025}.

To address this limitation, we propose a Bayesian-optimised weighted ensemble for cross-institution few-shot 3D abdominal and pelvic MRI segmentation. Multiple few-shot segmentation models are first adapted to a target anatomy using a small labelled support set. Bayesian optimisation is then applied on a separate small validation set to identify ensemble weights that maximise segmentation performance. The resulting weights are used to combine voxel-level probability maps through weighted averaging, producing a task-specific ensemble tailored to the target anatomy. By learning ensemble weights directly from limited target-task data, the proposed framework adapts to variations in anatomy, image appearance, institutional characteristics, and model reliability without requiring extensive manual tuning. From a deployment perspective, the framework provides a practical mechanism for incorporating data from previously unseen institutions into an existing segmentation workflow using only a small number of labelled cases, avoiding both costly retraining and manual selection of the most appropriate few-shot learning strategy. In this way, the proposed approach addresses a common data-engineering challenge in medical AI, which is adapting segmentation systems to new clinical sites where labelled data are scarce and imaging characteristics differ from those observed during development.

The contributions of this work are summarised: 1) we propose a Bayesian-optimised weighted ensemble framework that automatically adapts the contribution of multiple complementary few-shot segmentation algorithms to a target anatomy and institutional domain using only a small number of labelled examples; 2) we present a practical adaptation pipeline for deploying segmentation models to previously unseen institutions under severe annotation constraints, enabling automatic model selection and ensemble weighting without additional retraining; 3) we evaluate the proposed framework on a real multi-institution male pelvic MRI dataset and demonstrate statistically significant improvements over individual few-shot methods, fixed-weight ensembles, and training-from-scratch baselines under simultaneous label scarcity and institutional domain shift.

\section{Methods}

\begin{figure}
    \centering
    \includegraphics[width=0.95\linewidth]{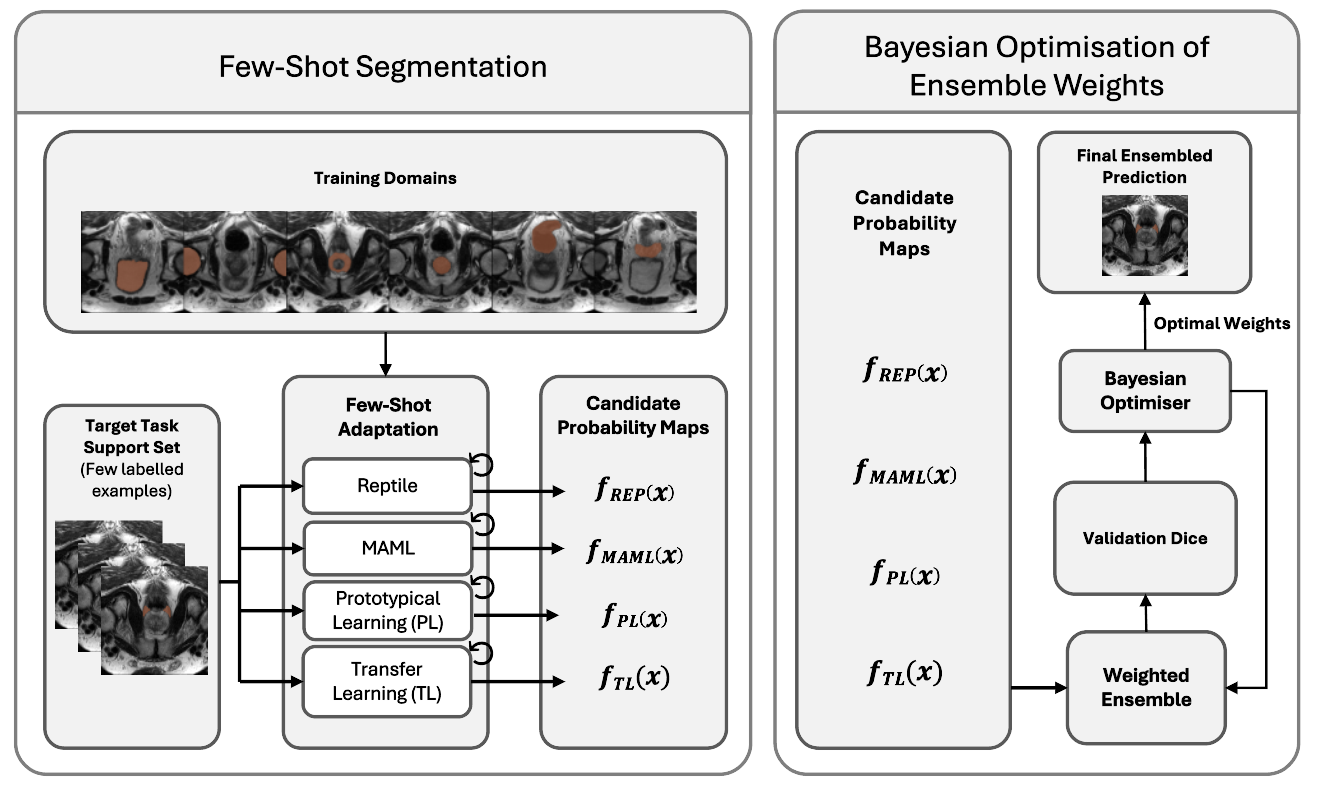}
    \caption{Overview of the proposed Bayesian optimisation of few-shot segmentation ensemble weights. }
    \label{fig:methods}
\end{figure}

\subsection{Few-shot segmentation}

Let $\mathcal{D}=\{\mathcal{D}_i\}_{i=1}^{S}$ denote a collection of source domains used to learn transferable segmentation representations, with each domain consisting of $\mathcal{D}_i=\{ (x_{i,j},y_{i,j})\}_{j=1}^{N_i}$, where $x_{i,j} \in \mathbb{R}^{H\times W\times D}$ is an image and $y_{i,j} \in \{0,1\}^{H\times W\times D}$ is the corresponding segmentation label. The source domains may correspond to different anatomical structures, institutions, or segmentation tasks.

Given a segmentation model $f(\cdot;\theta)$ with parameters $\theta$, few-shot learning seeks to learn transferable representations from the source domains that can be rapidly adapted to a previously unseen target domain. During source-domain training (referred to as meta-training in meta-learning approaches), model parameters are optimised across all source domains by minimising a segmentation loss $\mathcal{L}$:

\begin{equation}\label{eq:source_opt}
\theta^* = \arg\min_{\theta}
\sum_{i=1}^{S}
\mathbb{E}_{(x,y)\sim\mathcal{D}_i}
\left[
\mathcal{L}
\left(
f(x;\theta),y
\right)
\right]
\end{equation}

At deployment, a previously unseen target domain $\mathcal{D}_T$ is available. A small labelled support set $\{x_{T,j},y_{T,j}\}_{j=1}^{K}\sim\mathcal{D}_T$, where $K$ is typically small, is sampled from the target domain and used for adaptation. The source-domain model is adapted to the target domain using the support set, by minimising:

\begin{equation}
\theta^T = \arg\min_{\theta}
\frac{1}{K}\sum_{j=1}^K
\left[
\mathcal{L}
\big(
f(x_{T,j};\theta),y_{T,j}
\big)
\right]    
\end{equation}

The adapted model $f(\cdot;\theta^{T})$ is subsequently applied to previously unseen query images from the target domain to generate segmentation probability maps. Different few-shot learning approaches perform this adaptation through different mechanisms, including parameter fine-tuning, gradient-based meta-learning, and support-set feature matching. In this work, we treat the few-shot learner as a generic adaptation framework and make no assumptions regarding the specific adaptation strategy employed.

\subsection{Weighted ensembling}

Let $\{f_m(\cdot; \theta^{m,T})\}_{m=1}^{M}$ denote a collection of $M$ few-shot segmentation algorithms, each parameterised by $\theta^{m,T}$. Given a target-domain image $x_{T,j}$, the $m$-th model generates a voxel-wise probability map $f_m(x_{T,j}; \theta^{m,T})$.

The final ensemble prediction is obtained by combining the probability maps from all models using a weighted average: 

\begin{equation}\label{eq:ensembling}
f_{\mathrm{ens}}(x_{T,j})
=
\sum_{m=1}^{M}
w_m \times
f_m(x_{T,j}; \theta^{m,T}),
\end{equation}

subject to $w_m \geq 0$ and $\sum_{m=1}^{M} w_m = 1$, where $w_m$ denotes the contribution of the $m$-th few-shot learner.

\subsection{Bayesian optimisation of ensemble weights}

The optimal weight vector $\mathbf{w}=\{w_m\}_{m=1}^{M}$ is learned using a small labelled validation set from the target domain. Specifically, Bayesian optimisation is used to identify the weights that maximise segmentation performance:

\begin{equation}
\mathbf{w}^{*}
=
\arg\max_{\mathbf{w}}
g(\mathbf{w}),
\end{equation}

where $g(\mathbf{w})$ denotes the segmentation performance (e.g., Dice score) of the corresponding ensemble on the target-domain validation set. 

This optimisation problem is solved using Bayesian optimisation, which iteratively evaluates candidate weight vectors to efficiently identify high-performing ensemble configurations. The resulting optimal weights $\mathbf{w}^{*}$ are subsequently used to generate predictions on previously unseen query images from the target domain, following equation \ref{eq:ensembling}.

\section{Experiments}

\subsection{Dataset}

The data used in this work comes from the Cross-institution Male Pelvic Structures dataset \cite{li2023prototypical,li_2022_7013610}, with a total of 589 3D T2-weighted MR images. The eight structures segmented within this dataset are: bladder, bone, obturator internus, transition zone, central gland, rectum, seminal vesicle and neurovascular bundle. The neurovascular bundle and obturator internus were reserved as heldout structures for evaluation and the rest were used for training or monitoring. The heldout structures were chosen due to the relatively low prevalence of annotation of these structures in routine practice, to mimic genuine data scarcity. Out of the seven available institutions, institutions 3 and 4 were heldout from training for evaluation purposes, and the rest were used for training or monitoring. To summarise, evaluation is performed on the neurovascular bundle (NB) and obturator internus (OI) from institutions 3 and 4 with 74 and 82 samples respectively, with subgroup analyses presented in the results. Out of these samples, 8 were used to adapt the models and 8 were used to compute metrics to tune weightings in Bayesian optimisation, and the rest were used for evaluation.

\subsection{Network architectures} 

All few-shot learning algorithms used a U-Net backbone. The network followed an encoder-decoder architecture with four downsampling stages, a bottleneck, and four upsampling stages with symmetric skip connections. Each convolutional block contained two 3D convolutional layers followed by batch normalisation and ReLU activation. Downsampling was performed using $(2\times2\times2)$ max-pooling, and upsampling was performed using transposed convolutions. The final convolution was followed by a sigmoid activation to obtain voxel-wise foreground probabilities. Models were trained using a combined soft Dice and binary cross-entropy (BCE) loss.

\subsection{Few-shot learning algorithms}

The few-shot learning algorithms used in this work are: 1) Reptile \cite{nichol2018first}; 2) Model-Agnostic Meta-Learning (MAML) \cite{Finn2017}; 3) Prototypical Learning (PL) \cite{li2023prototypical}; and 4) Transfer Learning (TL), all pre-trained on the non-heldout structures and institutions. The models took approximately 16, 21, 23, 14 hours to train, respectively, using two Nvidia Ada 5000 GPUs, with training conducted in parallel. The inference times were 320 milliseconds on average on the same hardware. 

\subsection{Comparisons}

Our Bayesian adaptively-weighted ensemble (3 random initial calls plus 10 additional calls) is compared with: 1) equal-weighted ensemble; 2) random-search weighted ensemble (16 random calls); 3) grid-search weighted ensemble (16 combinations); 4)-7) individual few-shot algorithms; and 8)-10) recent state-of-the-art approaches. The computational budgets / numbers of calls are set based on values reported in recent literature. The comparisons are in terms of Dice score and corresponding standard deviation measured across the subgroup or entire dataset as specified. Comparisons 1)-3) and 8)-10) serve as current common practice and state-of-the-art. The equal-weighted ensemble also serves as an ablation study, whereby the impact of weighted vs simple averaging can be evaluated. Additionally, comparisons against individual algorithms are also ablations that allow evaluation against individual components of the weighted ensemble. Paired t-tests are performed, with p-values reported, for any comparisons made. Our code is openly available at: \url{github.com/kushuu/few-shot-3d-image-segmentation}.

\section{Results}

\subsection{Comparisons}

The results are summarised in Table \ref{tab:results}. The observed ensemble weights align with the performance observed for each of the constituent algorithms within the ensemble, where PL outperformed the remaining constituents, with statistical significance (all p-values$<$0.025) and consistently had higher ensemble weights in all ensembling approaches. For example, the final ensemble weights for NB institution 3, for our Bayesian optimisation framework were: [0.34, 0.21, 0.38, 0.07] for [Reptile, MAML, PL, TL], respectively. For Random-Search and Grid-Search the weights were: [0.29, 0.18, 0.35, 0.18] and [0.32, 0.13, 0.30, 0.25], respectively.

Interestingly, the Equal weights ensemble showed inferior performance to the constituent PL algorithm, with statistical significance (p-value=0.043). However, the Random-Search and Grid-Search ensemble outperformed all constituent algorithms, with statistical significance (all p-values$<$0.034). Statistical significance was not observed for comparisons between Random-Search and Grid-Search (p-value=0.098). 

Our proposed ensembling strategy based on Bayesian optimisation outperformed all tested algorithms, including constituents, baselines and recent state-of-the-art, with statistical significance (all p-values$<$0.024). This indicates that our approach is able to effectively select ensemble weights that improve segmentation performance, given the same computational budget as other baselines and state-of-the-art approaches. The proposed method also maintained consistently strong performance across both held-out institutions, suggesting improved robustness to institutional domain shift.

Interestingly, the ensemble weights were not distributed uniformly across constituent algorithms. Prototypical Learning (PL), which achieved the strongest individual performance, received the largest weight in all weighted ensemble approaches. However, Bayesian optimisation did not simply select the best-performing constituent. Instead, it assigned non-zero contributions to all methods, suggesting that each algorithm captured complementary information that improved overall segmentation performance when combined. These findings support the hypothesis that different few-shot learning approaches exhibit distinct strengths across anatomical structures and institutional domains, motivating task-adaptive rather than fixed-weight ensembling.

\begin{table}[!ht]
\centering
\begin{tabular}{|c|c|c|c|c|c|}
\hline
Structure & NB & OI & NB & OI & Aggregate \\
\hline
Institution & 3 & 3 & 4 & 4 & Aggregate\\
\hline
Method & \multicolumn{5}{c|}{}\\
\hline
\textbf{Ours}            & 0.567$\pm$0.076 & 0.561$\pm$0.083 & 0.542$\pm$0.092 & 0.556$\pm$0.091 & 0.557$\pm$0.081\\
\hline
Equal & 0.543$\pm$0.081 & 0.536$\pm$0.087 & 0.521$\pm$0.098 & 0.531$\pm$0.094 & 0.533$\pm$0.086\\
Random-Search & 0.552$\pm$0.077 & 0.547$\pm$0.084 & 0.529$\pm$0.094 & 0.537$\pm$0.093 & 0.541$\pm$0.082\\
Grid-Search & 0.553$\pm$0.078 & 0.548$\pm$0.084 & 0.531$\pm$0.094 & 0.541$\pm$0.092 & 0.543$\pm$0.084\\
\hline
Reptile & 0.520$\pm$0.091 & 0.531$\pm$0.095 & 0.502$\pm$0.103 & 0.515$\pm$0.099 & 0.516$\pm$0.089\\
MAML & 0.533$\pm$0.085 & 0.539$\pm$0.091 & 0.514$\pm$0.098 & 0.524$\pm$0.095 & 0.526$\pm$0.087\\
PL & 0.549$\pm$0.080 & 0.543$\pm$0.089 & 0.519$\pm$0.097 & 0.535$\pm$0.092 & 0.537$\pm$0.084\\
TL & 0.536$\pm$0.078 & 0.535$\pm$0.088 & 0.526$\pm$0.092 & 0.530$\pm$0.097 & 0.531$\pm$0.082\\
\hline
\cite{dang2021weighted}
& 0.543$\pm$0.082 & 0.534$\pm$0.088 & 0.517$\pm$0.099 & 0.529$\pm$0.095 & 0.531$\pm$0.087\\
\cite{DANG2024} & 0.546$\pm$0.080 & 0.543$\pm$0.086 & 0.522$\pm$0.097 & 0.534$\pm$0.093 & 0.536$\pm$0.085\\
\cite{zhou2023adaptively}
& 0.550$\pm$0.079 & 0.543$\pm$0.085 & 0.524$\pm$0.096 & 0.536$\pm$0.092 & 0.538$\pm$0.084\\
\hline
\end{tabular}
\caption{Our Bayesian adaptively-weighted ensemble approach compared with other baselines and state-of-the-art approaches.}
\label{tab:results}
\end{table}

\subsection{Qualitative Evaluation}

\begin{figure}[!ht]
    \centering
    \includegraphics[width=0.95\linewidth]{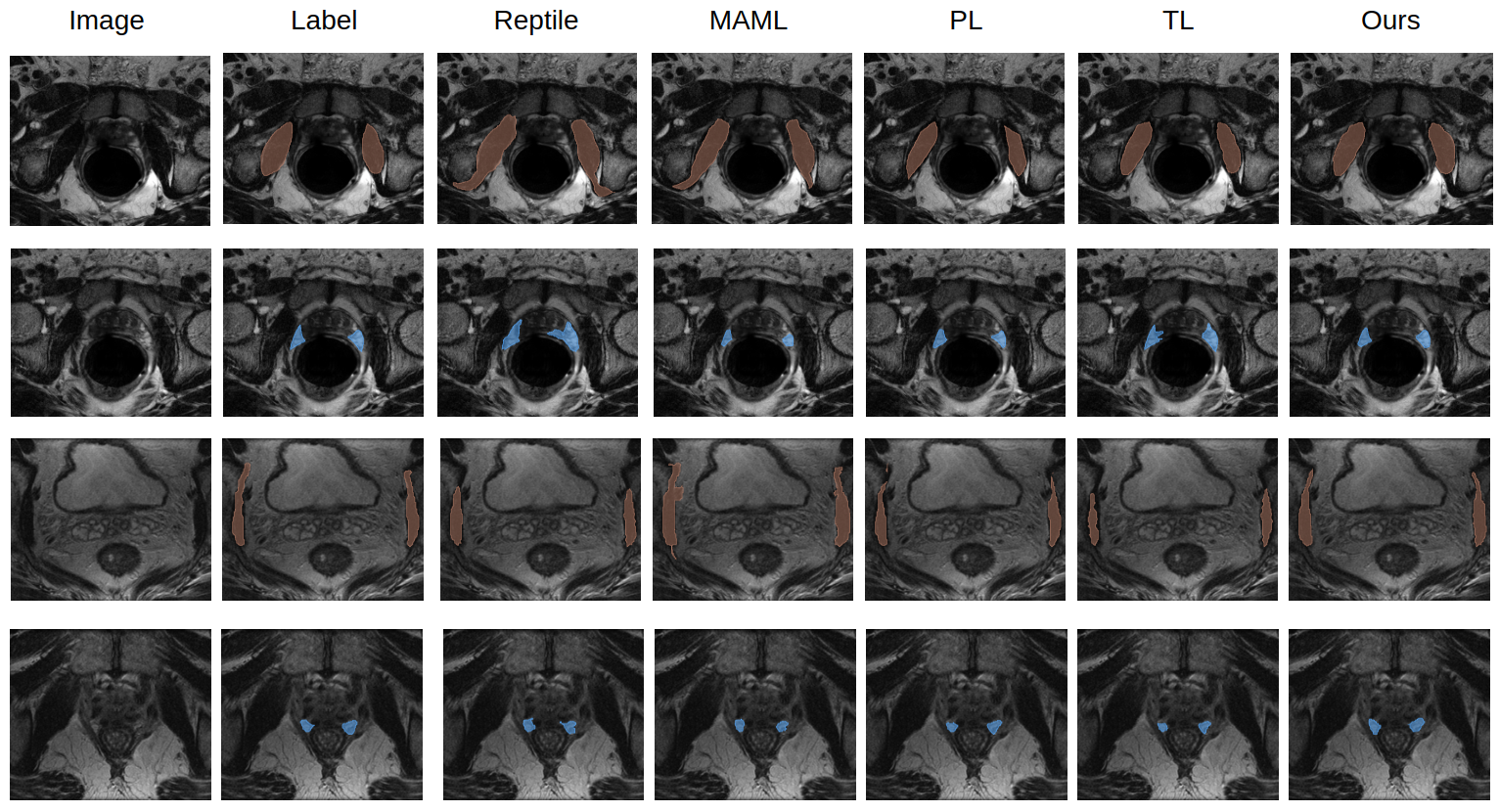}
    \caption{Qualitative comparisons of the ensembled prediction with its constituents and ground truth. Images are from institution 3 (top two rows) and institution 4 (bottom two rows), with the OI and NB segmented in red and blue, respectively.}
    \label{fig:qual_res}
\end{figure}

From the qualitative examples presented in Fig. \ref{fig:qual_res}, we can see that the our ensembled predictions are able to overcome the over- or under-segmentation of the constituents by combining their predictions. The difference in observed predictions between the constituents also indicates the variability in few-shot learning algorithms. For example, in institution 3 images Reptile can be seen consistently oversegmenting structures whereas in institution 4 it consistently undersegments. This variability highlights the need for task or institution level ensembling as presented in this work, as opposed to fixed ensembles.

\section{Conclusion}

We presented a Bayesian adaptively-weighted ensemble framework for few-shot anatomical segmentation under simultaneous label scarcity and institutional domain shift. By automatically learning task-specific ensemble weights from a small target-domain validation set, the proposed approach combines the complementary strengths of multiple few-shot learning algorithms without requiring manual model selection. Evaluation on the Cross-institution Male Pelvic Structures dataset demonstrated statistically significant improvements over individual few-shot learners, fixed-weight ensembles, training-from-scratch baselines, and recent state-of-the-art ensembling approaches. Analysis of the learned weights suggests that different few-shot methods capture complementary information across anatomical structures and institutions, motivating adaptive rather than fixed ensembling strategies. From a deployment perspective, while computationally expensive in development, the proposed framework provides a practical mechanism for incorporating new clinical sites into existing segmentation workflows using only a small number of labelled examples. Future work will investigate larger collections of constituent models, additional imaging modalities, and alternative optimisation strategies for learning ensemble weights.

\section*{Acknowledgements}

This work is supported by the International Alliance for Cancer Early Detection, an alliance between Cancer Research UK [EDDAPA-2024/100014] \& [C73666/A31378], Canary Center at Stanford University, the University of Cambridge, OHSU Knight Cancer Institute, University College London and the University of Manchester.

\newpage
\printbibliography

\end{document}